\documentclass[conference]{IEEEtran}

\usepackage{color,float,flushend,algorithm,algorithmic}
\usepackage{graphicx}
\usepackage[table]{xcolor}
\usepackage{mathrsfs}
\usepackage{tikz}
\usepackage{fancyheadings}
\newcommand*\circled[1]{\tikz[baseline=(char.base)]{\node[shape=circle,draw,inner sep=2pt] (char) {#1};}}
  
\begin{document}
\title{Learning Opposites with Evolving Rules}

\author{\IEEEauthorblockN{H.R.~Tizhoosh}
\IEEEauthorblockA{Centre for Pattern Analysis and Machine Intelligence\\
University of Waterloo\\ Waterloo, Ontario, Canada\\
Web: http://tizhoosh.uwaterloo.ca/}
\and
\IEEEauthorblockN{S.~Rahnamayan}
\IEEEauthorblockA{Electrical, Computer and Software Engineering\\
University of Ontario Institute of Technology\\ Oshawa, Ontario, Canada\\
Email: shahryar.rahnamayan@uoit.ca}}


\maketitle

\begin{abstract}
The idea of opposition-based learning was introduced 10 years ago. Since then a noteworthy group of researchers has used some notions of oppositeness to improve existing optimization and learning algorithms. Among others, evolutionary algorithms, reinforcement agents, and neural networks have been reportedly extended into their ``opposition-based'' version to become faster and/or more accurate. However, most works still use a simple notion of opposites, namely linear (or type-I) opposition, that for each $x\in[a,b]$ assigns its opposite as $\breve{x}_I=a+b-x$. This, of course, is a very naive estimate of the actual or true (non-linear) opposite $\breve{x}_{II}$, which has been called type-II opposite in literature. In absence of any knowledge about a function $y=f(\mathbf{x})$ that we need to approximate, there seems to be no alternative to the naivety of type-I opposition if one intents to utilize oppositional concepts. But the question is if we can receive some level of accuracy increase and time savings by using the naive opposite estimate $\breve{x}_I$ according to all reports in literature, what would we be able to gain, in terms of even higher accuracies and more reduction in computational complexity, if we would generate and employ \textbf{true opposites}? This work introduces an approach to approximate type-II opposites using evolving fuzzy rules when we first perform ``opposition mining''. We show with multiple examples that learning true opposites is possible when we mine the opposites from the training data to subsequently approximate $\breve{x}_{II}=f(\mathbf{x},y)$.     
\end{abstract}

\IEEEpeerreviewmaketitle

\section{What is the Problem?}
It was 10 years ago that ``\textbf{opposition-based learning}'' (OBL) was born. Since then, a modest but growing community of researchers has tried to use OBL to improve diverse optimization and learning techniques. Evolutionary algorithms, reinforcement agents, swarm-based methods, and neural networks, to mention a few, have been extended to employ oppositeness in their processing. Evaluating the error of weights and opposite weights, examining the rewards for actions and counter-actions, and examining the fitness of chromosomes and anti-chromosomes are examples for rethinking existing concepts by embedding opposite entities. Apparently, as a review of literature easily illustrates, using opposites can actually help accelerating many optimization and learning processes (see section \ref{sec:bg}). The main benefit of using opposites seems to be contributing to an accelerated convergence. Apparently by simultaneous consideration of entities and opposite entities, algorithms become capable of jumping over large portions of the solution landscape for difficult problems when they do not exhibit any significance.

Since the first paper was published on OBL in 2005 \cite{Tizhoosh2005a}, it has not been really clear how oppositeness can actually be captured with respect to the intrinsic behaviour of the problem at hand. The proposed scheme was that for any given $x\in [a,b]$ the opposite $\breve{x}$ can be given via $\breve{x}=a+b-x$. This, if at all, only makes sense for ``linear'' functions, in which case no particular algorithmic sophistication is required to begin with. But in spite of its simplicity, many works have reported benefits for using $\breve{x}=a+b-x$, which is called ``type I opposite'' in context of opposition-based computing \cite{Tizhoosh2008}.

After 10 years, some questions are still unanswered: Is there such thing as type II opposites? If they exist, how can we calculate them?  And most importantly, would using type II opposites bring any benefit to existing machine-learning and optimization methods? 

\section{The Idea}
\label{sec:idea}
What we call type II (or true) opposite of $x$, denoted with $\breve{x}_{II}$, is supposed to be meaningful for ``non-linear'' mappings and relationships, in contrast to type I opposites $\breve{x}^I$ that latently assume a linear relationship between in- and outputs.

Looking at the function $y=f(x_1,x_2,\dots,x_n)$ in a typical machine-learning or optimization context, one is generally fortunate to receive the output values $y$ for some input variables  $x_1,x_2,\dots,x_n$. However, the function $y=f(\cdot)$ itself is generally unknown otherwise there would be little justification for resorting to machine intelligence. Instead, one has some sort of evaluation function $g(\cdot)$ (error, reward, fitness etc.) that enables us to assess the quality of any guess  $\hat{x}_1,\hat{x}_2,\dots,\hat{x}_n$ delivering an estimate $\hat{y}$ of the true/desired output $y$.

The \textbf{idea} proposed in this paper is to use training data, whenever available, to perform \textbf{opposition mining} in order to approximate type II opposites gradually by learning the $x,y,\breve{x}_{II}$-relationship, or more precisely by learning $\breve{x}_{II}=f(x,y)$. Of course, if a large number of training data is available, then one could also just apply them \textit{at once}  instead of perpetual/continuous change. 

For every input $x$, in contrast to the type I opposites which are defined according to
\begin{equation}
\breve{x}_I=a+b-x,
\end{equation}
we propose to use the more meaningful type II opposites according to 
\begin{equation}
\breve{x}_{II}=\{ x_i | f(x_i) = y_{\min}+y_{\max} - f(x)\}. 
\end{equation}
There are several challenges for such an approach: \textbf{1)} the range of the output $y=f(\cdot)$ may not be a-priori known, hence we may need to continuously (in an evolving manner) update our knowledge of the output range $[y_{\min}, y_{\max}]$, \textbf{2)} the exact output for a type-II opposite may not be present in data. Hence, the approach should be capable of improving its estimation as data stream provides more clues about the gaps in our estimate of the in- and output relationship, and \textbf{3)} sufficient (representative) data may not be available at any given time; another reason in favour of an evolving approach.

In this work, we put forward a concise algorithm to learn opposites via fuzzy inference systems (FIS) with or without evolving rules depending on application whereas the evolving fuzzy systems may offer more versatility in many challenging applications (section \ref{sec:learnopp}). Before that, we briefly review the existing literature in section \ref{sec:bg}. Experiments and results, to verify the correctness and the usefulness of the proposed approach are described in section \ref{sec:exp}.     

\section{Reviewing Previous Works}
\label{sec:bg}
In this section, we briefly review the existing literature on opposition-based learning and the evolving fuzzy systems.
\subsection{Opposition-Based Learning}
Opposition-based learning was introduced in 2005 \cite{Tizhoosh2005a}. Motivated by finding an alternative for random initialization, it was observed that for many algorithms we start with a random guess and hope to move toward an existing solution in a speedy fashion. Examples were named such as the weights of a neural network, the population in genetic algorithms, and the action policy of reinforcement agents. Further, the paper stated that if we begin with a random guess far away from the existing solution, ``let say in worst case it is in the opposite location, then the approximation, search or optimization will take considerably more time, or in worst case becomes intractable''. The paper advocated that ``we should be looking in all directions simultaneously, or more concretely, in the opposite direction'' in order to have a higher chance of finding a solution in a shorter time, and established the following definition  \cite{Tizhoosh2005a}:

\textit{Opposition-Based Learning --} Let $f(x)$ be the function in focus and $g(\cdot)$ a proper evaluation function. If $x\in[a,b]$ is an initial (random) guess and $\breve{x}$ is its opposite value, then in every iteration we calculate $f(x)$ and $f(\breve{x})$. The learning continues with $x$ if $g(f(x))>g(f(\breve{x}))$, otherwise with $\breve{x}$. The evaluation function $g(\cdot)$, as a measure of optimality, compares the suitability of results (e.g. fitness function, reward and punishment, error function).

Of course, after many reports in the past 10 years we know that $x$ should not necessarily be a random guess at the beginning, but it can be a calculated or estimated value by a learning or optimization algorithm at any stage of the process. As well, we may not necessarily need to evaluate $g(f(x))$ and $g(f(\breve{x}))$ in all iterations.

Rahnamayan et al. introduced opposition-based differential evolution (ODE) which seems to be one of the most successful applications of OBL via integration within an existing algorithm \cite{Rahnamayan2006}-\cite{Rahnamayan2008b}.  Among other early works, there are papers on the application of OBL for reinforcement learning \cite{Tizhoosh2005b, Tizhoosh2006} where ``opposite actions'' were defined for specific problems, hence, not engaging in question of type-I versus type-II opposites. Defining opposite fuzzy sets, which may be mistaken with ``antonyms'' and their applications have been introduced too \cite{Fares2010b,Tizhoosh2009a, Tizhoosh2009b}. Tizhoosh and Ventresca \cite{Tizhoosh2008} presented a complete collection of oppositional concepts in machine intelligence in an edited volume, in which the term ``type-II opposition'' was coined in the chapter \emph{opposition-based computing}. Ventresca and Tizhoosh also extended simulated annealing \cite{Ventresca2007a} and introduced the notion of opposite transfer functions for neural networks \cite{Ventresca2007b}. 

Surveys are available to provide an overview of OBL methods. For instance, Al-Qunaieer et al. \cite{Fares2010a} provide a compact survey of oppositional methods. The most comprehensive survey of opposition-based methods so far has been published by Xu et al. \cite{Xu2014}.

Type II opposites have not been examined in-depth. However, as we will briefly review them in the following two paragraphs, recently two papers have independently proposed different possibilities for estimating type II opposites. 

Mahootchi et al. \cite{Mahootchi2014} provided some new ideas for type-II opposites when dealing with reinforcement learning methods such as Q- and \emph{sarsa} learning. Looking at actions $a$ that an agent could take to manipulate its environment, the type-I opposite action $\breve{a}$ was originally defined as `directional' opposite for grid-world agents, hence ``up'' was the opposite of ``down'', for instance. However, in a stochastic environment it makes sense to examine the discounted accumulated rewards $Q(s,a)$ for each state $s$ and estimate the opposite as follows: 
\begin{equation}
\breve{a}\! \in\! \left\{ \hat{a} | Q^t(i,\hat{a})\!\approx\! \max_b Q^t(i,b)\! +\! \min_b Q^t(i,b)\! -\!Q^t(i,a)\right\}
\end{equation}
To obtain new knowledge about the opposites from observations, the authors train an MLP network which is updated every several episodes. For higher accuracy, the approximate function could be updated online when the agent can interact with the stochastic environment. ``In other words, after the first training of the network, two sets of action-value functions exist, with the first being used to find the action that the agent should take and the approximate one being used to extract the opposites. This means that there is a mutual relationship between the agent and the opposite agent. This process may assist each agent to accomplish its tasks more efficiently in terms of convergence time and accuracy in real-world applications'' \cite{Mahootchi2014}. The authors also reported that MLP may not be efficient for large applications due to its computational expense. They suggest that one may substitute MLP with other types of function approximations such as fuzzy inference systems. 

Salehinejad et al. \cite{Salehinejad2014}, examining a type-II extension of opposition-based differential evolution (ODE) also point to the fact that the true opposites should be calculated via $\breve{f}(x) = y_{\min} + y_{\max} - f(x)$ to focus on the output but they use centroid-based method instead of depending on $\min$ and $\max$ specially where the landscape boundaries are unknown. A look-up table can help to find $\breve{f}(x)$, and if not present in the table, its value can be estimated via interpolation. In their algorithm, they use type-I and type-II opposition simultaneously. 

\subsection{Evolving Fuzzy Systems}
An \emph{evolving fuzzy system} has been initially introduced as an unsupervised method of updating the rule-based structure of a fuzzy inference system (FIS) in a non-iterative way \cite{angelov2001, angelov2002}. The main idea is that FIS can be extended as more data becomes available. The fuzzy rule base evolves (is extended) by adding more rules to a basic rule set formed by the initial fuzzy model or by replacing existing rules with ones that approximate the data better. When data are captured online, its potential is calculated, and the potentials of existing cluster centers are recursively updated. The potential of the new data is compared with the existing centers and one of the following decisions is made: 1) Replace one or more existing clusters with the new data point if its potential is higher than a certain threshold and the new data point is close to an old center, 2) Add the new data as a new cluster if its potential is higher than a certain threshold. 

The term ``evolving'' may imply that instead of retraining the initial system, the system should be updated online when new data become available. However, the evolution of rules in a fuzzy inference system can occur in a variety of ways, and as long as the initial fuzzy model (rule set) is continuously updated, no preference exists with respect to how the actual update is performed. In \cite{angelov2004}, the Takagi--Sugeno (T-S) type of fuzzy controller with an evolving structure was proposed for online identification of T-S systems. The fuzzy rules of this controller are expanded through the use of data collected during the control process, and is, thus, trained in a non-iterative (recursive) way. There have been approaches that use an iterative approach specially for the application where keeping and re-training the data does not pose a challenge \cite{Othman2011}. More recent approaches to evolving fuzzy systems have been proposed \cite{angelov2008},\cite{angelov2008b}. Evolving rules have been used for modelling nonlinear dynamic systems \cite{barros2007} and image classification and segmentation \cite{lughofer2010}\cite{Othman2011}\cite{Othman2014}.

Comprehensive treatment and reports on new advances in evolving fuzzy systems can be found in literature \cite{Lughofer2011,Lughofer2015,Pratama2014,Lughofer2015b}. There are other learning approaches one could use. For instance, participatory learning \cite{Yager1990} where compatibility between observations and beliefs plays an important role. 

In this paper, we propose a general, simple and efficient approach to learn type-II opposites using fuzzy inference systems. The emphasis is on \textbf{opposition mining} (finding opposites or quasi-opposites in the available training data) and \textbf{evolving rules} (which perpetually adjust the approximation of the mined opposites).

\section{Learning Opposites via Evolving Rules}
\label{sec:learnopp}
A fuzzy inference system (FIS) generally consists of a set of IF--THEN rules of the following form:

IF $x_1$ is $A_1$ AND $x_2$ is $A_2$ AND $\cdots$ AND $x_n$ is $A_n$\\
THEN $y$ is $B$

where $x_i, y \in X$ are variables defined in corresponding universes of discourse $X_i$ and $Y$, respectively, and $A_i$ and $B$ are fuzzy (sub)sets (in this work we investigate the problems with many inputs and one output only). These rules can be defined by experts. However, in most real-world applications they can be extracted from available data via clustering. A Takagi-Sugeno, or T--S fuzzy inference system operates with fuzzy rules of the following general form \cite{Takagi1985}:

IF $x_1$ is $A_1$ AND $x_2$ is $A_2$ AND $\cdots$ AND $x_n$ is $A_n$\\
THEN $y = f_j (x_1, x_2, \dots, x_n ), j = 1, 2, \cdots,N$

where $x_i$ and $y$ are variables defined in corresponding universes of discourse $X_i$ and $Y$, respectively, and $A_i$ is a fuzzy (sub)set. The function $f_j (x_1, x_2, \cdots, x_n $) is a crisp (nonfuzzy) function of $xi$. In general, the function $f_j$ is defined as the weighted combination of all variables $f_j (x_1, x_2, \cdots, x_n) = w-0 + w_1x_1 + w_2x_2 + \cdots + w_nx_n$.

The output is then calculated by

\begin{equation}
y = \frac{\sum_{j=1}^N f_j (x_1, x_2, \cdots, x_n) \mathcal{T}_{i=1}^{m_j}\mu_j(x_i)}{\sum_{j=1}^N  \mathcal{T}_{i=1}^{m_j}\mu_j(x_i)}
\end{equation}

where $N$ is the number of fuzzy rules, $n$ is the number of inputs (features), $\mu_j$ is the membership value of the $i$th input $x_i$ for the $j$th rule, $1 \leq m_j \leq n$, and $\mathcal{T}$ is a T-norm representing the logical conjunction.

In order to learn the type II (true) opposites with evolving fuzzy rules, we first have to sample the problem at hand to find existing (quasi-)opposites. This is the \textbf{opposition mining} stage that provides the data for clustering and the subsequent rule extraction. One generally assumes that the more data is available the better the approximation becomes because we expect that more data increases the probability for opposition mining to find more representative samples for rule generation in order to more accurately approximate $\breve{x}_{II}=f(x,y)$. We assume that the range of input variables is known, $x_i\in[x_{\min}^i, x_{\max}^i]$, but the range of output, $y_j\in[y_{\min}^j, y_{\max}^j]$, may be a-priori unknown. Since we are looking for the true (type-II) opposites, we need to calculate the (quasi-)opposite of the output for each sample. Subsequently, we find the closest value to that point and select its corresponding input as the opposite of the given input. If the inflow of data continues, the reliability of the mined opposites as training data increases which would result in higher accuracies for the approximated true opposites. 

\begin{figure}[htb]
\center
\includegraphics[width=\columnwidth]{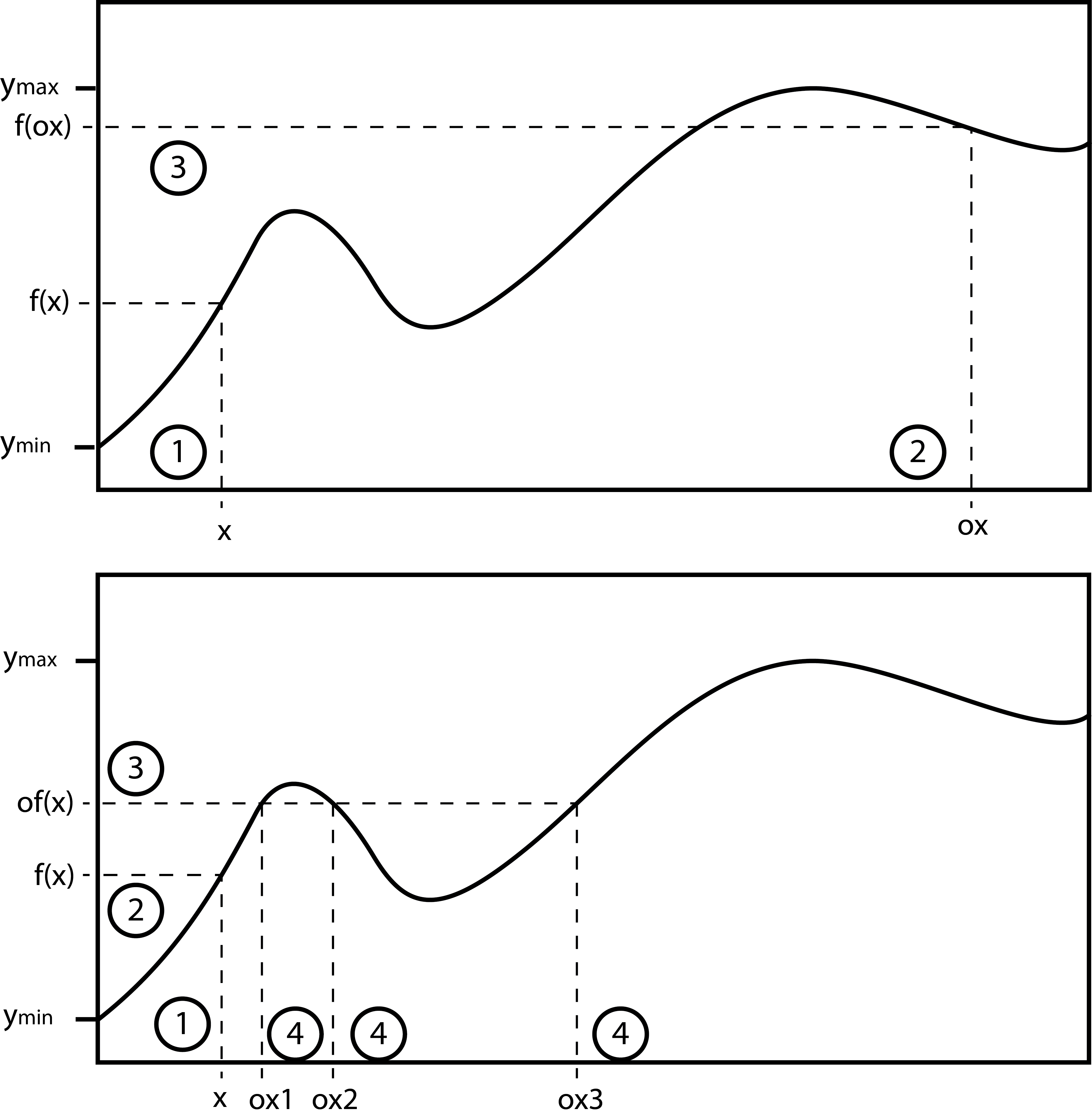}
\caption{Type I versus Type II Opposition. TOP: 1) variable x is given, 2) opposite of x, xo, is calculated via a+b-x . BOTTOM: 1) variable x is given, 2) f(x) is calculated, 3) opposite of f(x), namely of(x) is calculated, 4) opposite(s) of x are found: ox1,ox2 and ox3. }
\label{fig:typeIvstype2}
\end{figure}

As preparation for designing the learning procedure, let us look at Figure \ref{fig:typeIvstype2}:
\begin{itemize}
\item Calculating type I opposites is straightforward (top diagram). For any given $x$ we calculate its opposite $\breve{x}=a+b-x$ (denoted with ox on the x-axis) which can then be used to estimate the value of $f(\breve{x})$ (denoted with f(ox) on the y-axis). However, as it is apparent from the diagram, calculation of ox is completely detached from the output.
\item Type II opposites, in contrast, are based on a more realistic (or more intuitive) understanding of oppositeness (bottom diagram in Figure \ref{fig:typeIvstype2}). For any given $x$ ({\tiny\circled{1}}), we first need an estimate or evaluation of $f(x)$ ({\tiny\circled{2}}). Then, we find the opposite of $f(x)$, namely $\breve{f}(x)$ or of(x) in Figure \ref{fig:typeIvstype2} ({\tiny\circled{3}}). Any input that can produce outputs like of(x) is the type II opposite of $x$ (ox1, ox2 and ox3 on the x-axis) ({\tiny\circled{4}}). Of course, for non-monotonic functions, we may get multiple opposites for each input. In most cases, one of those opposites might suffice to exploit the potential benefits of OBL when validated with the available objective function.
\end{itemize}

Algorithm \ref{alg:genfis} describes the steps how we learn the opposites based on the same steps in the bottom diagram of Figure \ref{fig:typeIvstype2}. The approach consists of two distinct stages:

\textbf{Opposition Mining} (Lines 7 to 24) -- The training data is sampled to established the output boundaries. Depending on a specific oppositeness scheme, all data points are then processed to find the (quasi-)opposite of each input and the corresponding input $\breve{x}_{II}$, which is the type-II opposite. At the end of the opposition mining we have $n_s$ opposites. 

We also look at different ways of calculating opposition in general. Given any function $y=f(x)$ with variables $x\in [x_{\min},  x_{\max}]$ with a sample average of $\bar{x}$, one may calculate oppositeness for the input domain in different ways \cite{Tizhoosh2005a, Tizhoosh2008,Rahnamayan2014center, Rahnamayan2014cec} (\% denotes the modulo operator): 
 
 \begin{eqnarray}
T_1: \qquad    \breve{x}_I &=& x_{\min} + x_{\max} - x \\
T_2:\qquad     \breve{x}_I &=& \left(x + \frac{x_{\min} + x_{\max}}{2}\right)~\%~ x_{\max} \\
T_3: \qquad    \breve{x}_I &=& 2\bar{x} - x
 \end{eqnarray}           
 Analogously, given the output $y\in [y_{\min}, y_{\max}]$ with an average $\bar{y}$, one may calculate its opposite $\breve{y}_I$ in different ways: 
  \begin{eqnarray}
T_1: \qquad    \breve{y} &=& y_{\min} + y_{\max} - y \\
T_2:\qquad     \breve{y} &=& \left(y + \frac{y_{\min} + y_{\max}}{2}\right)~\%~ y_{\max} \\
T_3: \qquad    \breve{y} &=& 2\bar{y} - y
 \end{eqnarray}  
 
For $T_3$ scheme, we may go out of the existing boundary of the variables sometimes. For the experiments, we solved this problem by switching to scheme $T_1$ whenever necessary.  It is paramount to emphasize that any opposite calculate based on $\breve{y}$ is a type-II opposite.
 
\textbf{Learning the Opposites} (Lines 26 to 30) -- Here any learning and approximation method may be used to employ the results of opposition mining. As for our approach, namely the fuzzy inference systems (FIS), in- and outputs are clustered to extract rules. Performing fuzzy inference with these rules will then approximate the type-II opposites for new (unseen) inputs. 

In following section, we report the results of some experiments to verify the superiority of type II over type I opposites.  

\begin{algorithm}[htb]
\caption{Learning the opposites via Evolving Rules}
\begin{algorithmic}[1]
\label{alg:genfis}
\STATE -------------- Initialization -----------------
\STATE Set FIS mechanism, e.g. Takagi-Sugeno 
\STATE Determine the number of samples $n_s$
\STATE Set the number of clusters $n_c$
\STATE Select oppositeness scheme $T_i, i\in\{1,2,3\}$
\STATE -------------- \textbf{Opposition Mining} -----------------
\STATE Get sample points $<x_1^i,x_2^i,\dots,y^i>$ ($i=1,2,\dots,n_s$)
\STATE Calculate $y_{\min}$, $y_{\max}$, and $\bar{y}$
\FOR{$i=1:n_s$}
    \STATE $thisY \leftarrow y(i)$;
    \STATE Use $T_i$ to estimate the opposite of the ouput: 
    \STATE $\quad T_i=T_1$: $oppY \leftarrow y_{\min}+y_{\max}- thisY$
    \STATE $\quad T_i=T_2$: $oppY \leftarrow \left(thisY + \frac{(y_{\min}+y_{\max})}{2}\right)~ \%~ y_{\max}$
    \STATE	$\quad T_i=T_3$:	$oppY \leftarrow 2 \bar{y} - thisY$
    \STATE $\mathit{minDiff} \leftarrow \infty$ (a large number)
    \FOR{$j=1:n_s$}
        \STATE $thatY \leftarrow y(j)$
        \STATE $\mathit{thisDiff} \leftarrow | oppY - thatY |$
        \IF{$\mathit{thisDiff} < \mathit{minDiff}$}
            \STATE $\mathit{minDiff} \leftarrow \mathit{thisDiff}$\vspace{0.05in}
            \STATE $\breve{x}_{II}(i) \leftarrow x(j)$ 
        \ENDIF
    \ENDFOR
\ENDFOR
\STATE ----------- Train FIS --------------
\STATE Set $<x_1^i,x_2^i,\dots,y^i>$ as input \vspace{0.05in}
\STATE Set $<\breve{x}_{1,II}^i,\breve{x}_{2,II}^i,\dots>$ as output \vspace{0.05in}
\STATE Partition the data into $n_c$ clusters
\STATE Extract fuzzy rules 
\STATE Save the FIS data
\end{algorithmic}
\end{algorithm}

\section{Experiments and Results}
\label{sec:exp}
We conduct multiple experiments to demonstrate the usefulness of opposition mining and learning opposites with evolving fuzzy rules. We examine 9 simple functions with known inverse functions to verify the correctness of the algorithm, and to establish a better understanding of the approach. Using functions with known inverse relations is a very reliable testing vehicle because the true opposites can be easily calculated for validation. As well, we employed Matlab\textsuperscript{TM} ``\textbf{genfis3}'' function to implement a clustering-based Takagi-Sugeno fuzzy inference system (note: Matlab\textsuperscript{TM} ``\textbf{genfis3}'' was trained incrementally to achieve the rule evolution). We experimented with number of clusters $n_c$ to be 30 or 60 clusters (with relatively similar results) and set the number of maximum iterations to maxIter$=5000$. The fuzzy exponent was set to $m=2$ and the clustering error threshold was fixed at $\epsilon = 0.00001$. 

\subsection{First Experiment Series}
\label{sec:pexp}
\begin{figure}[htb]
\center
\includegraphics[width=0.6\columnwidth]{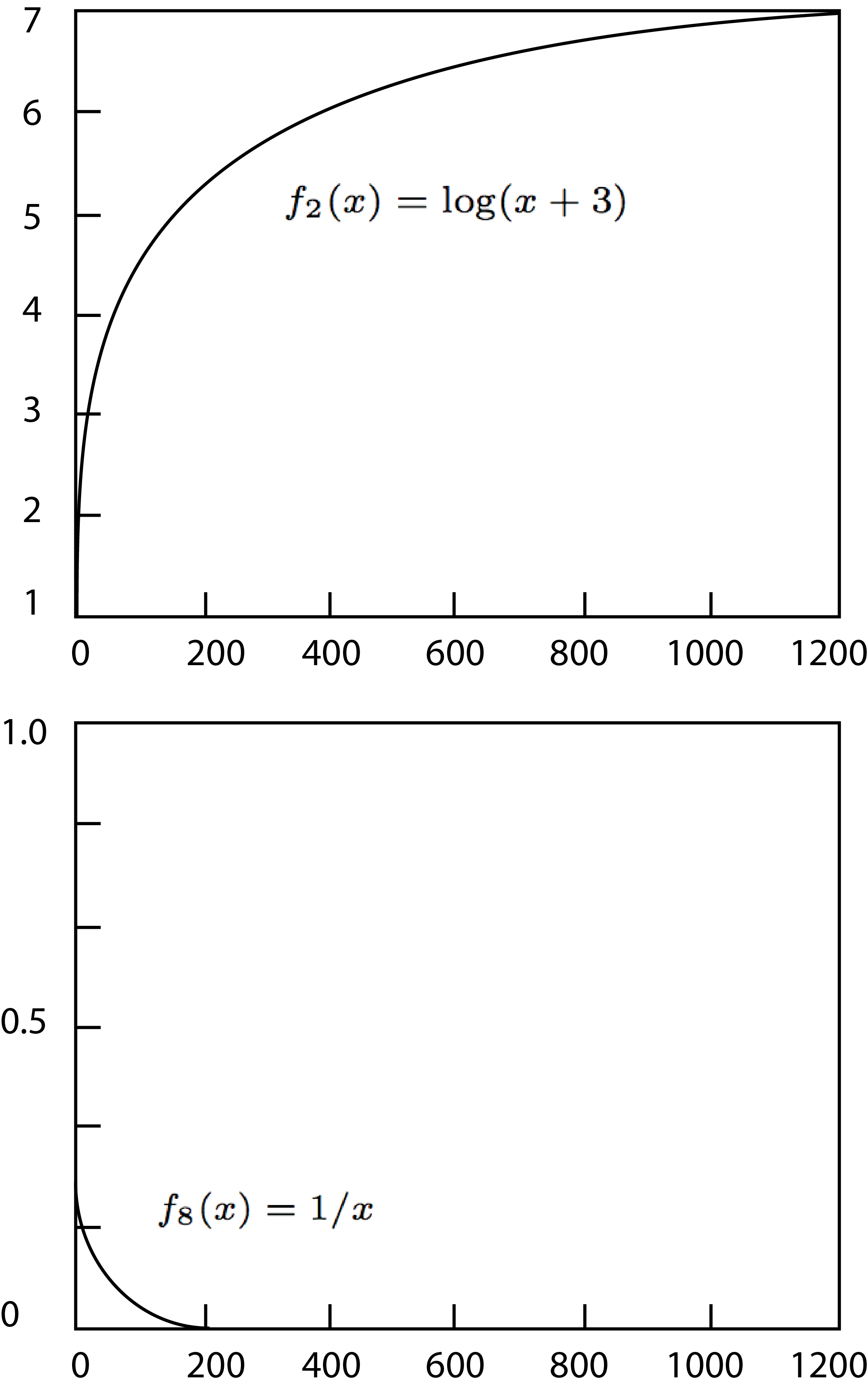}
\caption{Examples for the test functions: $f_2(x)$ (top) and $f_8(x)$ (bottom) according to equations provided in Table \ref{table:shahryarResults}. The simplicity of the functions allows us to calculate the inverse (hence the true opposites) but they are still difficult enough to demonstrate the limitations of the type-I opposition as large errors in Table \ref{table:shahryarResults} demonstrate.}
\label{fig:allfunctionsdiag}
\end{figure}

The results for 9 functions are summarized in Table \ref{table:shahryarResults} (also see Figure \ref{fig:allfunctionsdiag}). The simplicity of the functions enable us to easily formulate the inverse function such that we can exactly calculate the \textbf{true opposites}. With exception of $f_7(x)$, the inverse of all other test functions are given in Table \ref{table:shahryarResults}: 

\begin{eqnarray}
\label{eq:f9inv}
f_7^{-1}(x)\! &=&\! \frac{1}{9\left(\frac{y}{2} + \left(\left(\frac{y}{2} - \frac{29}{54}\right)^2\!-\!\frac{1}{729}\right)^\frac{1}{2}\! -\! \frac{29}{54}\right)^\frac{1}{3} } \\ 
&+& \left(\frac{y}{2} + \sqrt{\left(\left(\frac{y}{2} - \frac{29}{54}\right)^2 - \frac{1}{729}\right)} - \frac{29}{54}\right)^\frac{1}{3} - \frac{1}{3} \nonumber
\end{eqnarray}

For every function 100 random samples were drawn in the opposition mining stage. The error for $\breve{x}_I$ and $\breve{x}_{II}$ is calculated through direct comparison with the true opposite $\breve{x}^*$ available through inverse functions: error$(\breve{x}_{I})=|\breve{x}^*-\breve{x}_{I}|$ and error$(\breve{x}_{II})=|\breve{x}^*-\breve{x}_{II}|$. 

\textbf{Analysis of the Results --} 
\begin{itemize}
\item Type-II opposite is clearly more accurate than type-I opposite with exception of linear function $f_3(x)$.
\item The opposition scheme $T_1$ seems to be the best representation of oppositeness. $T_2$ is the best oppositeness scheme only for $f_8(x)$.
\item If we only examine type I opposites, $T_3$ is better than $T_1$ and $T_2$ for 7 out of 9 test functions. 
\end{itemize}	

\begin{table*}[htb]
\caption{The results for test functions (best results highlighted): For every function both Type I and Type II opposites are calculated using the three different oppositeness scheme. The average error $m$ and its standard deviation $\sigma$ is calculated using direct comparison with the known true opposite via inverse function.}
\begin{center}
\begin{tabular}{|l|c|c|c|}
		& Oppositeness	& Type-I Opposites & Type-II Opposites \\
Function and its inverse  & scheme & $m\pm\sigma$ for 30 runs & $m\pm\sigma$ for 30 runs \\ \hline\hline
 $f_1(x) = (2x + 8)^3$ 	& $T_1$ & $378.91\pm 178.48$ & \cellcolor[gray]{0.8} $\mathbf{22.62\pm 	24.01}$ \\
 $f_1^{-1}(x) = (y^{1/3} - 8)/2$	& $T_2$ & $337.20\pm 303.62$ & $78.12\pm 96.86$ \\
 				& $T_3$ & $234.27\pm 187.97$ & $155.25\pm 	228.39$ \\ \hline\hline
 $f_2(x) = \log(x+3)$		& $T_1$ & $478.06\pm 270.07$ & $ \cellcolor[gray]{0.8}\mathbf{18.95\pm 29.47}$ \\
 $f_2^{-1}(x) = \exp(y)-3$		& $T_2$ & $470.85\pm 296.33$ & $24.00\pm 51.03$ \\
				& $T_3$ & $529.12\pm 1881.12$ & $557.06\pm 2051.47$ \\ \hline\hline	 
 $f_3(x)= 2x$	 & $T_1$ & $ \cellcolor[gray]{0.8}\mathbf{0\pm 0}$ &	$0.02\pm 0.01$ \\
 $f_3^{-1}(x) = y/2$			& $T_2$ & $ \cellcolor[gray]{0.8}\mathbf{0\pm 0}$ &	$210.66\pm 137.20$ \\
				& $T_3$ & $0.51\pm 2.90$ & $0.82 \pm 0.53$ \\ 		\hline\hline
 $f_4(x)=x^2$			& $T_1$ & $285.91\pm 117.22$ & $ \cellcolor[gray]{0.8}\mathbf{21.77\pm 19.91}$ \\
 $f_4^{-1}(x) = \sqrt{y}$		& $T_2$ & $248.87\pm 297.11$ & $118.34\pm 129.50$ \\
				& $T_3$ & $152.22\pm 129.68$ & $120.17\pm 185.21$\\ \hline\hline
 $f_5(x) = \sqrt{x}$		& $T_1$ & $313.05\pm 140.76$ & $ \cellcolor[gray]{0.8}\mathbf{0.028\pm 0.018}$ \\
 $f_5^{-1}(x) = y^2$			& $T_2$ & $317.94\pm 335.52$ & $161.85\pm 156.19$ \\
 				& $T_3$ & $96.79\pm 99.41$ & $101.09\pm 156.69$ \\ \hline\hline
 $f_6(x) = x^{3/2}$		& $T_1$ & $183.72\pm 72.06$ & $ \cellcolor[gray]{0.8}\mathbf{15.38\pm 13.35}$ \\
 $f_6^{-1}(x) = y^{2/3}$		& $T_2$ & $171.61\pm 280.48$ & $150.75\pm 142.32$ \\
				& $T_3$ & $83.11\pm 73.16$ & $59.28\pm 98.56 $ \\ \hline\hline
 $f_7(x) = x^3 + x^2 + 1$	& $T_1$ & $380.63\pm 177.82$ & $ \cellcolor[gray]{0.8}\mathbf{22.98\pm 26.77}$ \\ 
 (for $f_7^{-1}(x)$ see Eq.\ref{eq:f9inv})	& $T_2$ & $328.18\pm 297.96$ & $75.55\pm 93.71$ \\
				& $T_3$ & $131.57\pm 85.13$ & $130.22\pm 177.85$ \\ \hline\hline
 $f_8(x) = 1/x$			& $T_1$ & $503.26\pm 292.20$ & $3.46\pm 13.49$ \\
 $f_8^{-1}(x) = 1/y$			& $T_2$ & $496.60\pm 289.02$ & $ \cellcolor[gray]{0.8}\mathbf{0.56\pm 0.27}$ \\
				& $T_3$ & $434.52\pm 338.03$ & $39.72\pm 163.92$ \\ \hline\hline
 $f_9(x) = \sqrt{(x+1)}/3$	& $T_1$ & $306.25\pm 138.12$ & $ \cellcolor[gray]{0.8}\mathbf{0.03\pm 0.02}$ \\ 
 $f_9^{-1}(x) = 9y^2-1$		& $T_2$ & $299.37\pm 324.00$ & $159.08\pm 165.29$ \\
				& $T_3$ & $95.63\pm 92.46$ & $97.60\pm 147.43$ \\ \hline
\end{tabular}
\end{center}
\label{table:shahryarResults}
\end{table*}%

\subsection{The Second Experiments Series}
In the second experiment series, we looked at the effect of evolving rules: What happens if data is fed into the system one-by-one or block-wise? Intuitively, one expects that through clustering of more data, a more compact rule base can be generated which can more accurately approximate the true opposites. 

\begin{figure*}[htb]
\begin{center}
\includegraphics[width=0.32\textwidth]{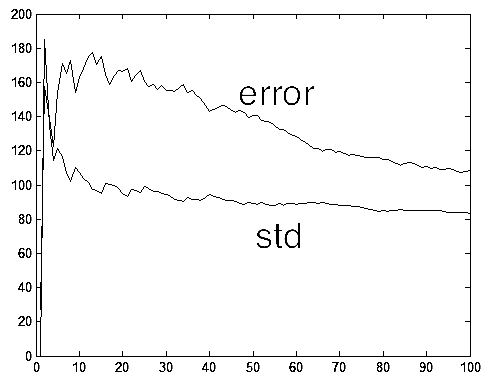}
\includegraphics[width=0.32\textwidth]{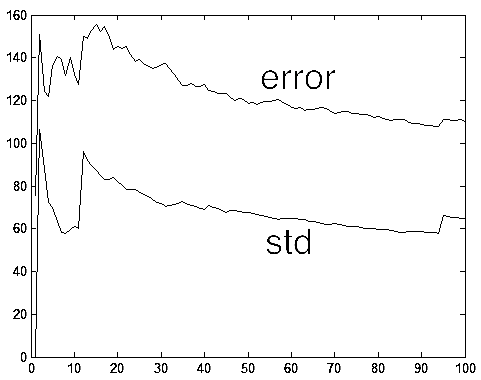}
\includegraphics[width=0.32\textwidth]{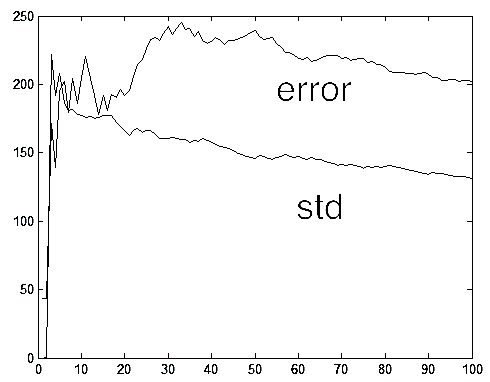}
\caption{The errors and standard deviations (std) for three different runs: In each run opposites are first trained offline with 100 samples (not shown here). The diagrams show the error of estimating the true opposites based on $f_1(x)$ (Table \ref{table:shahryarResults}) with 100 new opposites online whereas the rules are clustered and evolved with $101, 102,\cdots, 199, 200$ samples, respectively.}
\label{fig:threediag}
\end{center}
\end{figure*}

\begin{figure*}[htb]
\center
\includegraphics[width=0.6\textwidth]{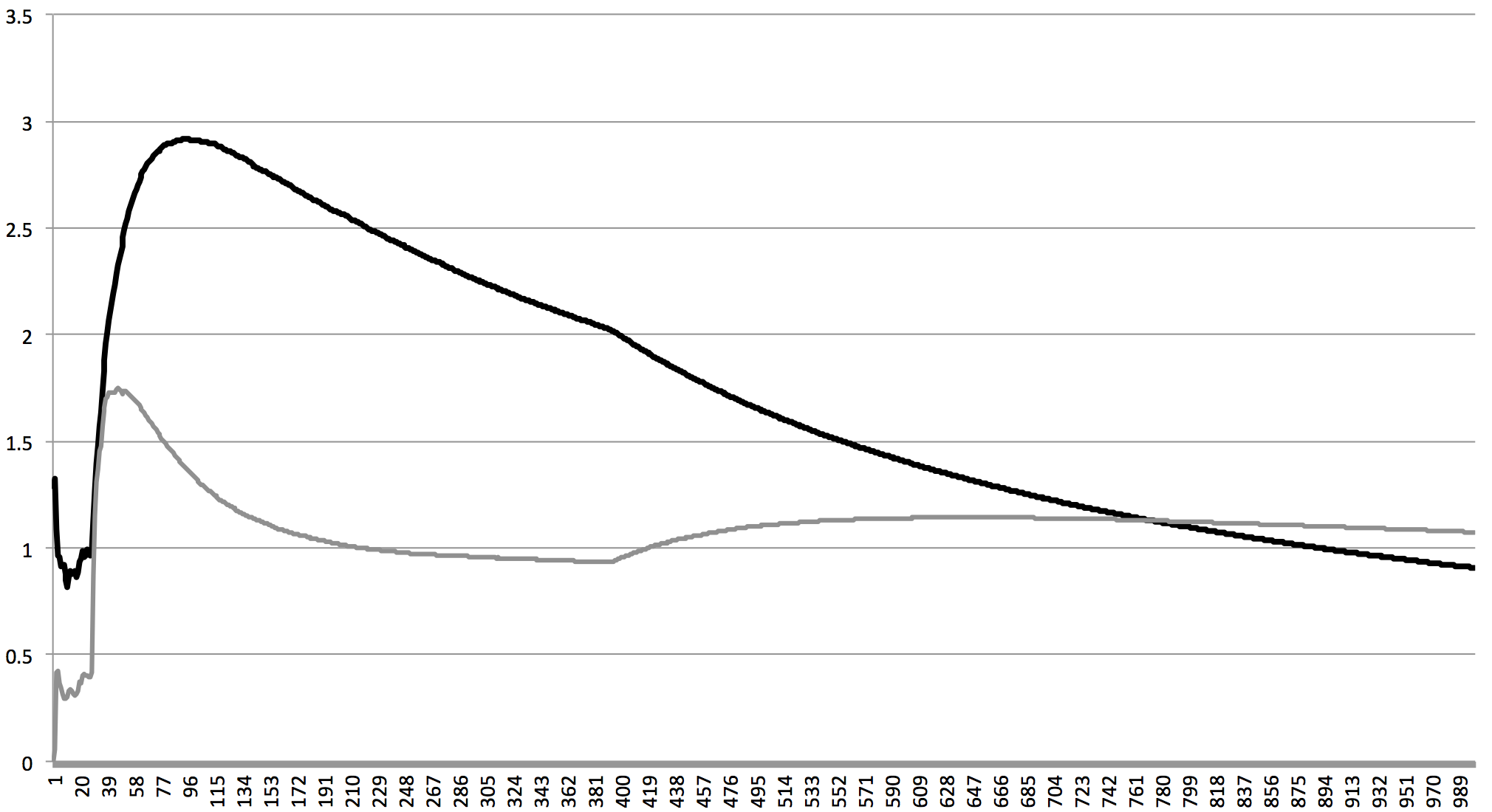}
\caption{The evolving opposites gradually become more accurate after processing each additional data point. The average error (black) and standard deviation (gray) decrease with each new sample. At the beginning there may be fluctuations due to lack of sufficient number of data to extract reasonable clusters and hence meaningful rules. Standard deviation may grow larger than mean if a small sample contains variables with large range.}
\label{fig:f3EvolvingError}
\end{figure*}

As Figure \ref{fig:threediag} shows some representative examples, the average error and standard deviation of approximated opposites decrease over time as the estimated values by evolving rules become more and more representative of the true opposites. We also observed a typical \emph{pinnacle point} at which the error reaches its maximum when one observes the evolving opposites long enough (Figure \ref{fig:f3EvolvingError}). This appear to be due to the intrinsic nature of evolving steps as at the beginning there are not enough data points to extract \emph{good} rules, thus resulting in large errors.  

\subsection{Third Experiment Series}
\label{sec:opexp}
In this section, we test 3 standard optimization functions which are commonly used in the literature of global optimization. For our convenience and without any loss of generality, we use only 2D functions, two variables $x_1$ and $x_2$. We learn the opposites with $n_s=1000$ samples and then test the function optimization with $0.1\times n_s$ new samples. Two random variables are generated and used to call FIS for their opposites. The combination of these two may provide a lower error as seen for the examined functions. Now, as  we have established the superiority of evolved type II opposites over type I opposites in previous sections, here we also examine type II opposites in conjunction with random samples (also see \cite{RahnamayanRAND, RahnamayanDIST}). Here we would like to verify whether the fundamental statement of OBL is true: Simultaneous consideration of guess and opposite guess provides lower errors at each iteration of learning and optimization processes. 

Given a function $f(x_1,x_2)=0$ we generate two random samples $x_1^r$ and $x_2^r$ and we calculate the error. Then, we approximate the opposites  $\breve{x}_1^r$ and $\breve{x}_2^r$ and calculate the error. As generally proposed in OBL schemes, we can simultaneously look at all results and continue with the best results depending on the evaluation of the function call. That means we continue with the results of  $x_1^r$ and $x_2^r$ if Error$(f(x_1^r,x_2^r))$$<$Error$(f(\breve{x}_1^r,\breve{x}_2^r))$ otherwise with  $\breve{x}_1^r$ and $\breve{x}_2^r$. The opposites $\breve{x}_i^r$ can be type I (as used in existing literature) or type II (as we are proposing). The question is which one can provide a statistically significant benefit when used in conjunction with the initial random guess $x_i^r$.   

\textbf{Ackley Function --} The Ackley function is given as
\begin{eqnarray}
f(x_1,x_2)=20\left(1 - \exp{\left(-0.2 \sqrt{0.5 (x_1^2 + x_2^2)}\right)}\right)-\\ \nonumber
\exp{\left(0.5(\cos{(2\pi x_1)} + \cos{(2\pi x_2)})\right)} + \exp(1).
\label{eq:ackley}
\end{eqnarray}
 
The global minimum is $f(x_1, x_2) = f(3, 0.5) = 0$ where the search domain is $-35\!<\!x_i\!<\!35$.  The results for Ackley function are given in Table \ref{table:ackley}. The random guesses (first column) and their type I opposites (third column) were statistically the same (null hypothesis could not be rejected). Their simultaneous consideration therefore does not provide any benefit. In contrast, the type II opposites (second column) are not only different but also exhibit lower errors. 
            
\begin{table}[ht]
\caption{Errors for Ackely Function}
\begin{center}
\begin{tabular}{|c|c|c|c|}
&  $x_1^r$ and $x_2^r$ &  $\breve{x}_{1,II}^r$ and $\breve{x}_{2,II}^r$ &  $\breve{x}_{1,I}^r$ and $\breve{x}_{2,I}^r$\\ \hline
1. run	 & $284.89\pm 263.32$ 	& 	$10.35 \pm 9.21$	& $284.89 \pm263.32$ \\
2. run	 & $20.45\pm 2.39$ 		&	$2.09 \pm 2.09$ 	& $20.45 \pm 2.39$ \\
3. run 	& $20.61\pm 1.50$		&	$3.90 \pm 0.41$	& $20.61\pm 1.50$ \\
4. run	& $20.24\pm 2.27$		&	$2.40 \pm 1.19$	& $20.24 \pm 2.27$\\ 
5. run	& $20.63\pm 1.71$		&	$4.06\pm   1.01$	& $20.63 \pm  1.72$\\ \hline
\end{tabular}
\end{center}
\label{table:ackley}
\end{table}%

\textbf{Booth Function --} The Booth function is given as
\begin{equation}
f(x_1,x_2) =(x_1 + 2x_2 - 7)^2 + (2x_1 + x_2 - 5)^2.
\end{equation}

The global minimum is $f(x_1, x_2) = f(1, 3) = 0$ where the search domain is $-10\!<\!x_i\!<\!10$.  The results for Booth function are given in Table \ref{table:booth}. Apparently, the random guesses (first column) and their type I opposites (third column) are the same distribution. Their simultaneous consideration therefore does not provide any benefit. In contrast, the type II opposites (second column) are different. The simultaneous consideration of guess and opposite guess does provide a plausible benefit.

\begin{table}[ht]
\caption{Errors for Booth Function}
\begin{center}
\begin{tabular}{|c|c|c|c|}
&  $x_1^r$ and $x_2^r$ &  $\breve{x}_{1,II}^r$ and $\breve{x}_{2,II}^r$ &  $\breve{x}_{1,I}^r$ and $\breve{x}_{2,I}^r$\\ \hline
1. run	 & $448.78\pm 460.46$ 	& 	$1930 \pm 612.73$	& $432.54 \pm 444.90$ \\
2. run	 & $403.22\pm 425.78$ 	&	$2000 \pm 575.18$ 	& $400.88 \pm 424.10$ \\
3. run 	& $470.40\pm 556.03$	&	$1875 \pm 717.06$	& $374.51\pm 388.70$ \\
4. run	& $382.35\pm 405.19$	&	$2103 \pm 545.19$	& $400.07 \pm 465.64$\\ 
5. run	& $422.20\pm 483.02$	&	$2058\pm   609.94$	& $415.63 \pm  455.30$\\ \hline
\end{tabular}
\end{center}
\label{table:booth}
\end{table}%

\textbf{Bukin4 Function --} The Bukin4 function is given as
\begin{equation}
f(x_1,x_2)=100 \sqrt{|| x_2 - 0.01x_1^2||} + 0.01 ||x_1+10||.
\end{equation}

The global minimum is $f(x_1, x_2) = f(-10, 0) = 0$ where the search domain is $-15\!<\!x_1\!<\!-5$ and $-3\!<\!x_2\!<\! 3$. 
The results for Bukin4 function are given in Table \ref{table:bukin4}. Apparently, the random guesses (first column) and their type I opposites (third column) are the same distribution (null hypothesis could not be rejected). Their simultaneous consideration therefore does not provide any benefit. In contrast, the type II opposites (second column) are both different and deliver lower errors.

\begin{table}[ht]
\caption{Results for Bukin4 Function}
\begin{center}
\begin{tabular}{|c|c|c|c|}
&  $x_1^r$ and $x_2^r$ &  $\breve{x}_{1,II}^r$ and $\breve{x}_{2,II}^r$ &  $\breve{x}_{1,I}^r$ and $\breve{x}_{2,I}^r$\\ \hline
1. run	 & $290.14\pm 271.39$ 	& 	$1.75 \pm 1.59$	& $290.14 \pm 271.39$ \\
2. run	 & $317.69\pm 253.78$ 	&	$7.34 \pm 9.6$ 		& $317.69 \pm 253.78$ \\
3. run 	& $301.05\pm 301.05$	&	$1.26 \pm 1.03$	& $301.05\pm 289.75$ \\
4. run	& $299.18\pm 262.70$	&	$49.06 \pm 34.26$	& $299.18 \pm 262.70$\\ 
5. run	& $282.59\pm 274.73$	&	$6.43\pm   5.01$	& $282.59 \pm  274.73$\\ \hline		
\end{tabular}
\end{center}
\label{table:bukin4}
\end{table}%
As apparent from the three tables, the simultaneous consideration of random guess and its opposite, as already demonstrated in existing literature, has clear benefit as always one of them is much closer to the solution. This shows that the learned type-II opposites do satisfy the fundamental assumption of OBL, namely that the simultaneous consideration of a guess and its opposite delivers a shorter path to the solution.


\section{Summary}
Ten years after introducing opposition-based learning, there exists no general algorithm to generate opposites for learning and optimization purposes. In this paper, taking the 50th anniversary of fuzzy sets as a pretext, we introduced an approach to learn true (type II) opposites via evolving fuzzy inference systems. The core idea in this paper is  ``opposition mining'' to extract (quasi-)opposites by processing the available training data. This makes the learning of opposites possible as it extracts the necessary data (specially the desired outputs) for any learning scheme to approximate. The evolving fuzzy rules can then capitalize on these data points and refine the estimate of opposition with perpetual incorporation of future data points. We tested the proposed algorithm with simple test functions, examined its evolving aspect, and investigated three widely used optimization benchmark functions as well to verify both the correctness and usefulness of type II opposites. 

We used the Takagi-Sugeno fuzzy systems in this work to learn the true opposites. This approach does employ a t-norm and hence triggers axis-parallel rules. However, It has been clearly shown in recent literature that arbitrarily rotated rules are able to outperform the conventional axis-parallel ones \cite{Lughofer2015b}. This is certainly a possibility for potential improvement in future works. As well, in this work we only investigated the problems with one output. The extension to multiple output systems will certainly be of interest.

The authors are investigating the benefits of the proposed approach for complex learning and optimization problems, and hope to report the results in foreseeable future. 

\textbf{Acknowledgment --} The authors would like to thank the anonymous reviewers who provided us with critical feedback and pointed to relevant literature we had missed. The authors would also like to thank \textbf{NSERC} (Natural Sciences and Engineering Research Council of Canada) for providing funding for this project in form of two Discovery Grants. 

\bibliographystyle{siam}

\end{document}